\documentclass[runningheads]{llncs}
\usepackage[T1]{fontenc}
\usepackage{graphicx}
\usepackage{amssymb}
\usepackage{amsmath}
\usepackage{subcaption}
\usepackage{multirow}

\begin{document}
\title{Towards exploring adversarial learning for anomaly detection in complex driving scenes}
\titlerunning{Towards exploring adversarial learning for anomaly detection ...}
%
\author{Nour Habib\orcidID{0009-0007-6060-6177} \and
Yunsu Cho\orcidID{0009-0004-3052-040X} \and
Abhishek Buragohain\orcidID{0000-0002-9503-9498} \and
Andreas Rausch\orcidID{0000-0002-6850-6409} }
\authorrunning{N. Habib et al.}
%
\institute{Institute for Software and Systems Engineering, Technische Universit\"at Clausthal, Arnold-Sommerfeld-Straße 1, Clausthal-Zellerfeld 38678, Germany \\
\email{\{nour.habib, yunsu.cho, abhishek.buragohain, andreas.rausch\}@tu-clausthal.de}\\
\url{https://www.isse.tu-clausthal.de/}}
\maketitle              
\begin{abstract}
One of the many  Autonomous Systems (ASs), such as autonomous driving cars, performs various safety-critical functions. Many of these autonomous systems take advantage of Artificial Intelligence (AI) techniques to perceive their environment. But these perceiving components could not be formally verified, since, the accuracy of such AI-based components has a high dependency on the quality of training data. So Machine learning (ML) based anomaly detection, a technique to identify data that does not belong to the training data could be used as a safety measuring indicator during the development and operational time of such AI-based components. Adversarial learning, a sub-field of machine learning has proven its ability to detect anomalies in images and videos with impressive results on simple data sets. Therefore, in this work, we investigate and provide insight into the performance of such techniques on a highly complex driving scenes dataset called Berkeley DeepDrive.

\keywords{Adversarial Learning, Artificial Intelligence, Anomaly Detection, Berkeley DeepDrive(BDD).}
\end{abstract}

\section{Introduction}
\label{sec:introduction}

Autonomous systems have achieved tremendous success in various domains, such as autonomous cars, smart office systems, smart security systems, and surveillance systems. With such advancements, nowadays,  autonomous systems have become very common in our daily life, where we use such systems regularly even in various safety-critical application domains such as financial analysis. All these current developments in various autonomous systems are due to increased performance in the field of machine learning techniques. Many of these autonomous systems have been developed as hybrid systems combining classically engineered subsystems with various Artificial (AI) techniques. One such example is autonomous driving Vehicles. In autonomous vehicles, the trajectory planning subsystem is usually designed in classic engineered ways, whereas the perception part of such vehicles to understand the surrounding environment is based on AI techniques. Both these parts are combined, so they could perform as a single system to execute various safety-critical functions.\cite{Youtie.2017}

\begin{figure}
\centering
  \includegraphics[width=0.5\textwidth]{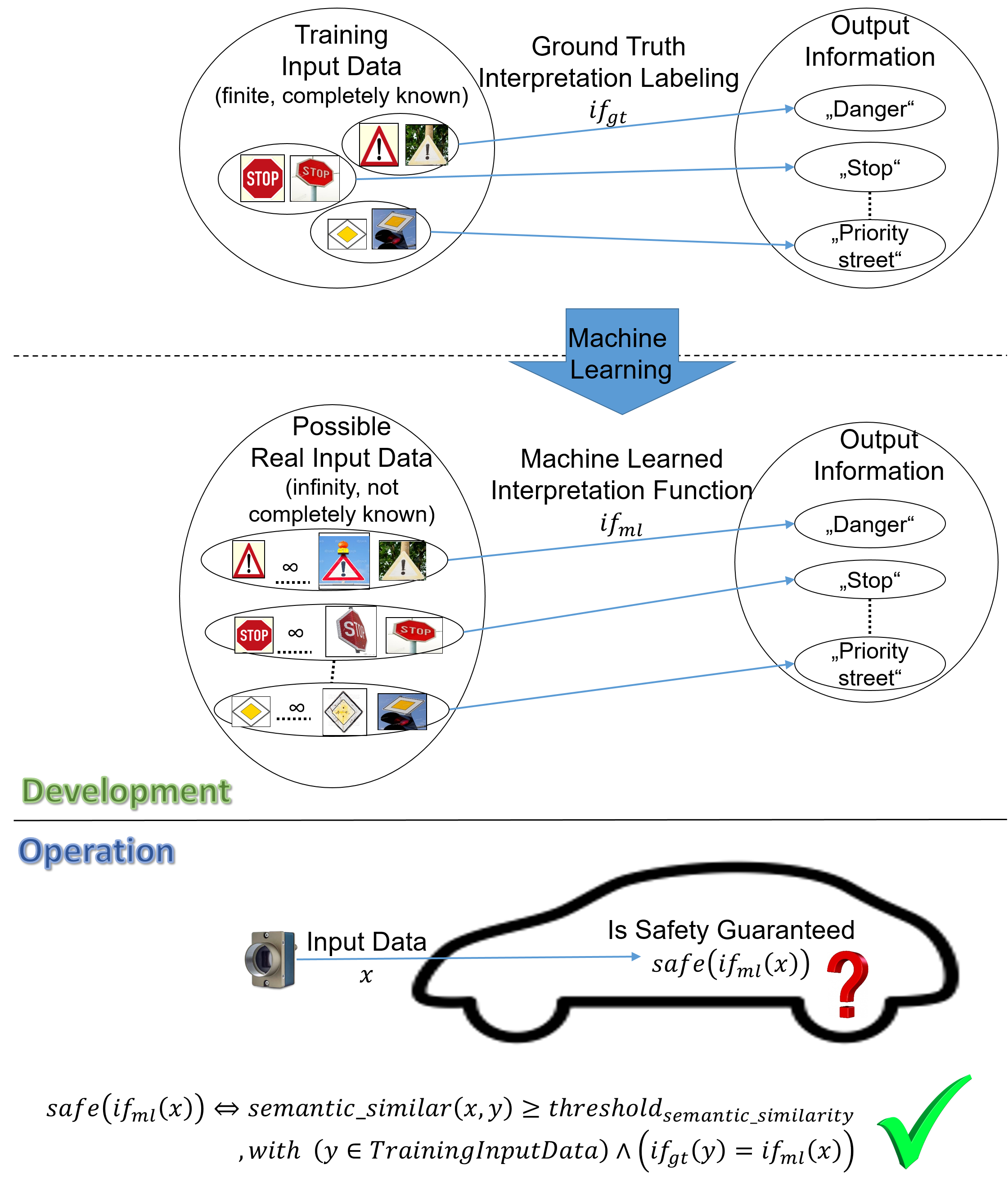}
  \caption{Operational time check of dependability requirements.\cite{Raulf.2021}} \label{fig:dependability_requirements}
\end{figure}

During the design and development phase of the perception subsystems in autonomous vehicles, first perceptions engineers label the training data. Then they use these labeled data and machine learning frameworks to train an interpreted function. To explain this concept of training an interpreted function, we can take the example of a perception task, where the trained interpreted function can perform the classification of traffic signal images into its correct traffic sign, as illustrated in Figure. \ref{fig:dependability_requirements}. In general training data is the mapping of a finite set of input data to its corresponding output information. In this case, input data can be compared to images of traffic signs and output information can be compared to its correct label which is traffic sign class. So once this machine-learned interpreted function is trained using this traffic sign training data, during the test time, it could process any image to one of the specified traffic sign classes in the training data's output information. However if we consider such machine learned interpreted function to be a part of a real Autonomous vehicle's perception subsystem stack, one important question arises, that is to what extent, the output $if_{ml} (x)$ of such interpreted function  $if_{ml}$ is reliable or safe enough,  so that other subsystems in the AVs can rely on.\cite{Rausch.2021}

The process behind the development of classical engineered systems vastly differs from AI-based systems\cite{Rushby.}. During the development process of a classical engineered system, first, semi-formal requirements specifications are created. Such requirement specifications are later used during the testing and verification process of the developed engineered system. But in the case of the development phase of AI-based systems, the creation of such requirement specifications is never done. And instead, it is replaced by a collection of training data. But such training data always does not contain all the information required and is sometimes incomplete \cite{Rausch.2021}. In other cases, due to inaccurate labeling, some amount of these training data consist of errors.,

The manufacturers of AI-based systems especially, in a domain like ASs, need to meet very high standards, so that they can satisfy the requirements of the regulations imposed on them. But the current engineering methods struggle to guarantee the dependability requirements from the perspective of safety  security, and privacy in a cost-effective manner. Because of such limitations,  the development engineers struggle to completely guarantees all the requirements of such AI-based systems during the development phase. Aniculaesei et al.\cite{Aniculaesei.2018} introduced the concept of a dependability cage, which can be used to test a system's dependability requirements both during the development and operational phase.

In the concept of a dependability cage as is shown in the Figure. \ref{fig:core}, the Quantitative monitor plays an important role to validate the output of the machine-learned interpreted functions. The concept of the Quantitative Monitor is such that, it tries to check if the sensor data  currently being configured  by the system as the input data for the ML-interpreted function is semantically similar enough to the ground truth of the training data, used during the development time of the machine learned interpreted  function. In this way, the Quantitative monitor tries to verify if the output information of the machine-learned interpreted function is correct and safe. If the current input data to the interpreted function is not semantically similar enough to the training data, then, it is an indication that the output information of this function is not safe or reliable to be used for making safety-critical tasks. Anomaly detection is a promising method, which is used to identify if the input test data $(x)$, is similar or not similar to the ground truth$(y)$ of training data based on some semantic similarity metric
$(semantic\_similar(x,y)\ge threshold_{semantic\_similarity})$(cf. Figure. \ref{fig:dependability_requirements}). With the evolution of the Artificial intelligence field, many promising methods for anomaly detection such as Generative Adversarial Networks,  Autoencoder,  and Variational Autoencoder have been experimented to  provide the measurement of semantic similarity between the test data and the training data.

The fundamental principle for using auto-encoder for the task of anomaly detection was introduced by  Japkowicz et al.\cite{japkowicz.1995}. Here the auto-encoder is first trained to minimize the distance error between the input image to the auto-encoder and its output, that is the reconstructed image. And once trained, during the test time, the auto-encoder takes a new image as input and  tries to produce a reconstructed image closer to the original input image. Now if the distance error between the input and the reconstructed output image is higher than a certain threshold, then the input image is considered to be an anomaly, or else, it is classified as not an anomaly. Amini et al.\cite{amini.2018} also demonstrated the same concept but with training data containing daylight driving scenes and anomaly images consisting of nighttime driving scenes.

But unlike Autoencoder approaches, many of Generative adversarial networks(GANs) based anomaly detection follows a slightly different underlying  principle for discriminating between anomaly and not anomaly data. Generative Adversarial Networks (GANs) have some advantages over Autoencoders and Variational Autoencoders. One of the main advantages of GANs for anomaly detection is their ability to generate realistic data samples \cite{SchleglSWSL17}. In a GAN, a generator network learns to create new samples that are similar to the training data, while a discriminator network learns to distinguish between real and fake samples\cite{SchleglSWSL17}. This means that a GAN can learn to generate highly realistic samples of normal data, which can be useful for detecting anomalous samples that differ significantly from the learned distribution\cite{SchleglSWSL17}.
Another advantage of GANs is that they can be trained to detect anomalies directly, without the need for a separate anomaly detection algorithm. This is achieved by training the discriminator network to not only distinguish between real and fake samples but also between normal and anomalous samples. This approach is known as adversarial anomaly detection and has been shown to be effective in detecting anomalies in a variety of domains\cite{SchleglSWSL17}.
 The anomaly detection using GANs-based techniques will further be  discussed in section \ref{sec:previousGAN} of the paper.

However, most of these papers have evaluated  their techniques on simple data sets such as MNIST \cite{LeCun.1999.}, CIFAR \cite{Krizhevsky2009}, UCSD \cite{Mahadevan.anomaly.2010}. Images in such data sets have very simple features, for instance, in MNIST, only one number is present per image. Similar is the case in CIFAR, where one object class is present per image. Another issue with these data sets is that they are of low resolution. Because in the real world, the driving scene images contain various class of objects, taken in various lighting conditions and weather such as raining day, night, day, etc. So the application of such anomaly detection techniques on complex driving scenes is still needed to be evaluated.

So to evaluate the performance of these GAN-based anomaly detection techniques on real-world driving scenes, we will first reproduce their work using their settings. Once we are able to reproduce their work, then we will evaluate their technique on a complex real-world driving scenes dataset of our choice. For the purpose of this evaluation, we have selected the BDD dataset \cite{Fisher.2020} as our driving scenes dataset. The reason behind using this dataset is, it has high variability in terms of class of objects, number of objects, weather conditions, and environment.  

The rest of the paper is organized as follows: In Section 2, we provide a brief introduction to dependability cage monitoring architecture for autonomous systems;  In Section \ref{sec:previousGAN}, we provide an overview of GAN-based novelty detection works; in Section 4, the research questions for our work are introduced; In Section 5, we present the description, concept, and dataset for the selected GAN technique; In Section 6, we present the evaluation of the selected technique; in Section 7, we provide a short summary of the contribution of our work and future work in the direction of GAN based technique for anomaly detection.

\section{Dependability Cage - A brief overview}

\begin{figure}
\centering
  \includegraphics[width=0.6\textwidth]{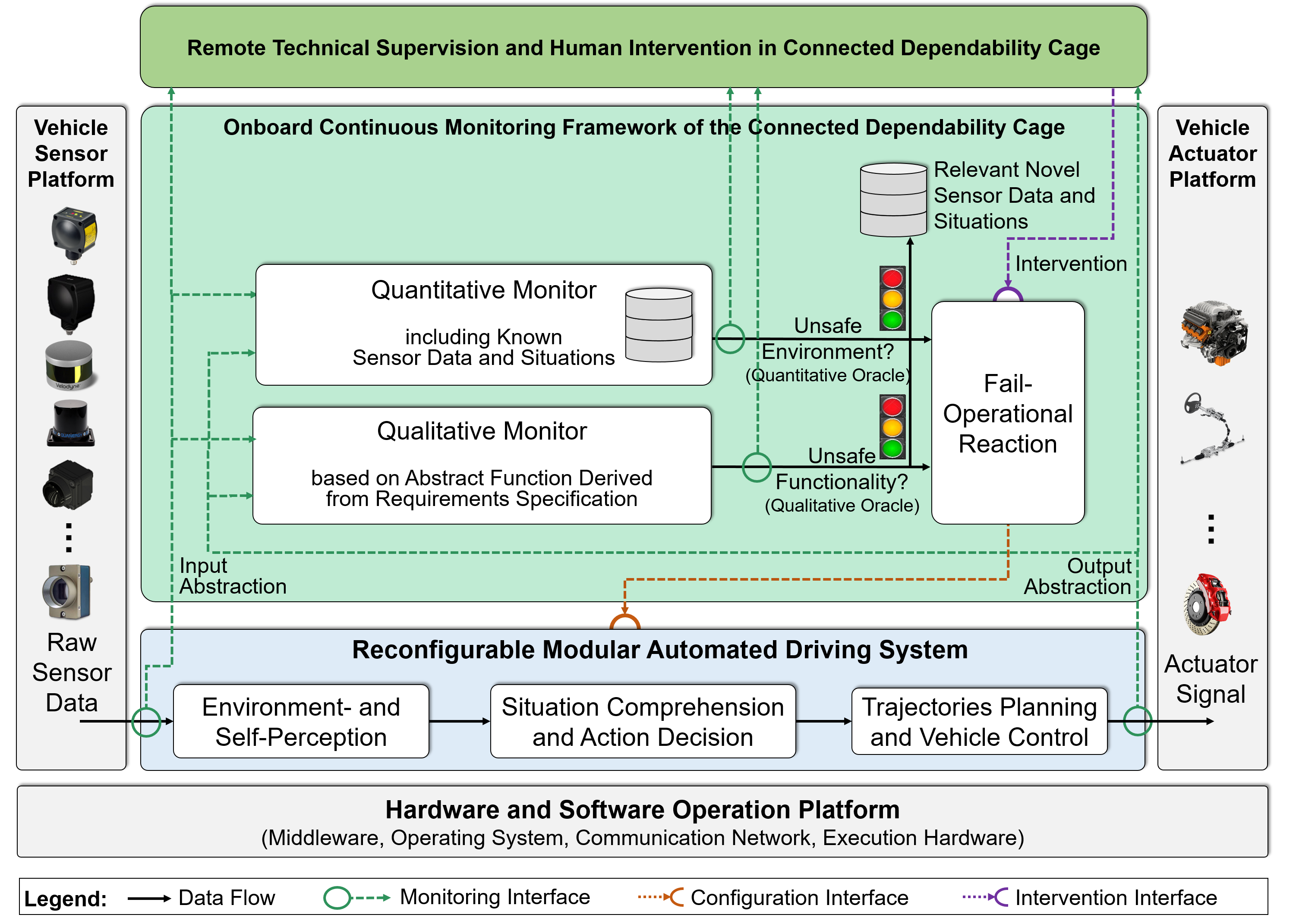}
  \caption{Continuous Monitoring framework in the dependability cage. \cite{Rausch.2021}} \label{fig:core}
\end{figure}

To overcome the challenges posed by engineering-dependent autonomous systems,  the concept of a dependability cage has been proposed in \cite{Aniculaesei.2018}\cite{Mauritz.2014}\cite{Behere.2015}\cite{Behere.2016}\cite{Maurer.2015}. These dependability cages are derived by the engineers from existing development artifacts. The fundamental concept behind these dependability cages is a continuous monitoring framework for various subsystems of an autonomous system  as shown in Figure \ref{fig:core}.

Firstly a high-level functional architecture of an Autonomous Vehicle(AV) has been established. It consists of three parts 1. environment self-perception, 2. situation comprehension, and action decision. 3. trajectory planning and vehicle control \cite{Behere.2016}\cite{Behere.2015}\cite{Maurer.2015}. The continuous monitoring framework resolves two issues that could arise. 1. shows  the system the correct behavior based on the dependability requirements. the component handling this issue is called the qualitative monitor. 2. Makes sure the system operates in a situation or environment that has been considered and tested during the development phase of the autonomous system. The component handling  this issue is called the quantitative monitor as shown in Figure \ref{fig:core}.

For the monitors to operate reliably, both of them require consistent and abstract data access from  the system under consideration. This is handled by the input abstraction and output abstraction components. In Figure \ref{fig:core}, these components  are shown as the  interfaces  between the autonomous systems and the two monitors, as monitoring interface. Both these input and output abstraction components convert the autonomous systems' data into an abstract representation based on certain user-defined interfaces. The type and data values of this representation are decided  based on the requirements specification and the dependability criteria derived by the engineers, during the development phase of the autonomous systems.\cite{Rausch.2021}

The quantitative monitor observes the abstract representation of the environment, it receives from the  autonomous system's  input and output abstraction components. For every situation, the monitors evaluate the abstract situation, in real-time, if it is known or tested during the development time of the autonomous system. The information about these tested situations is provided by a knowledge base to the quantitative monitor.

Since the study in this work is carried out taking quantitative monitoring into consideration, we will talk about quantitative monitoring in detail.  For a better understanding of the qualitative monitor, refer to the work of Rausch et al. \cite{Rausch.2021}. If one of the above-mentioned monitors detects any unsafe and incorrect behavior in the system, an appropriate safety decision has to be taken to uphold the dependability requirements. For this purpose, the continuous monitoring framework has a fail operation reaction component which receives the information from both the monitors and must bring the corrupted systems to a safe state. One such fail operational reaction could be a graceful degradation of the autonomous system as stated in \cite{Aniculaesei.2019}. As part of the safety decision, the system's data will be recorded automatically. These recorded data then can then be transferred back to the system development so that, they could be analyzed and  be used to resolve any unknown system's  faulty behaviors. 
	
Finally to realize such a Quantitative Monitor, an efficient solution to classify between known and unknown sensor data is required. As mentioned in the previous section, for this we will first, re-evaluate some of the state-of-the-art generative adversarial network (GAN) based anomaly detection techniques on data sets published in their work. Once we successfully complete the first step, we will evaluate their performance on a driving scenes dataset quite similar to scenes encountered by an autonomous vehicle in the real world. A study of these anomaly detection techniques and the selection of some of these techniques for the purpose of the evaluation will be described in the following sections.

\section{Previous works on GAN-based anomaly detection} \label{sec:previousGAN}

Anomaly detection has become a genuine concern in various fields, such as security, communication, and data analysis, which makes it a significant interest to researchers. In this paper, we wanted to explore its ability  to detect unknown  and known driving scenes from the perspective of an autonomous vehicle. Unknown driving scenes can be considered as anomalies that were not considered during the development time of the autonomous vehicle. Whereas known driving scenes are  detected as not anomaly, which was considered during the development time of the autonomous vehicle. Various theoretical and applied research have been published in the scope of detecting anomalies in the data. In the following subsections, we will review some of the research papers from which, we selected the approaches. These selected  approaches  will be our reference for further evaluation in this paper later on.

\subsection{Generative adversarial network-based novelty detection using minimized reconstruction error}

\begin{figure}
\centering
  \includegraphics[width=0.7\textwidth]{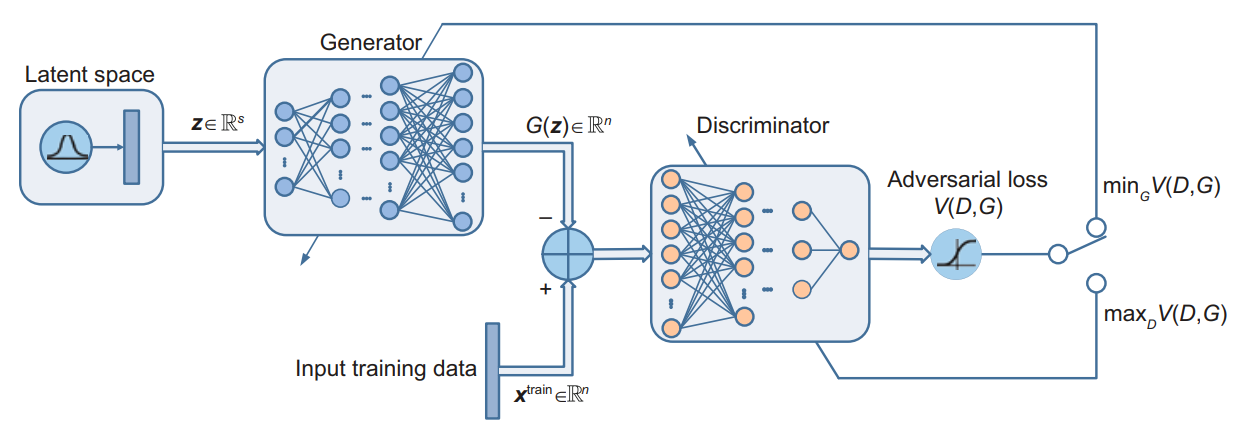}
  \caption{The structure of DC-GAN \cite{vanilladcgan}} \label{fig:dcganrf1}
\end{figure}

This paper provides an investigation of the traditional semi-supervised approach of deep convolutional generative adversarial networks (DC-GAN) for detecting the novelty in both the MNIST digit database and the Tennessee Eastman (TE) Dataset \cite{vanilladcgan}. Figure \ref{fig:dcganrf1} presents the structure of the network of DC-GAN that has been used in the paper. The generator used a random vector Z (Latent space) as input for generating samples G(Z), and the discriminator discriminates whether those samples belong to the Dataset distribution, i.e., indicating them as not-novel images, or they don’t belong to the Dataset distribution and they are indicated as novel images\cite{vanilladcgan}. Only data with normal classes were used during the training, so the discriminator learns the normal features of the dataset to discriminate the outlier in the data during the evaluation. The loss function of GAN that is used in the paper is minimax as presented in equation \ref{minimax1} \cite{vanilladcgan}.

\begin{equation}  \label{minimax1} 
\min_{G}\max_{D} V(D,G) =  E_{x\sim_{P_{\textrm{data}(x)}}}[log D(x)] +  E_{x\sim_{P_{\textrm{z}}(z)}}[log (1-D(G(z)))]
\end{equation}

The generator tries to minimize the error of this loss function and reconstruct better images G(z), while the discriminator D(X) tries to maximize the error of this loss function and indicate the generated samples as fake images \cite{vanilladcgan}. Figure \ref{fig:dcganrf1} presents the structure of the network of DC-GAN that has been used in the paper.

The discriminator classifies whether those samples belong to the dataset distribution, i.e., indicating them as not-novel images, or they don’t belong to the dataset distribution and they are indicated as novel images \cite{vanilladcgan}. Data with normal classes were used during the training so the discriminator learns the normal features of the dataset to discriminate the outlier in the data during the evaluation \cite{vanilladcgan}. The loss function of GAN that is used in the paper is the minimax loss function as presented in equation \ref{minimax1}. The evaluation metrics that are used for model evaluation of the MNIST dataset are the novelty score fg(x) or called the G-score, and the D-score fd(x)\cite{vanilladcgan}. The G-score is calculated as presented in equation \ref{fgequation} to minimize the reconstruction error between the generated sample G(z) and the reference actual image x. While the D-score is the discriminator output( decision), and it is varied between 0 (Fake - Novel) to 1 (real - not novel)\cite{vanilladcgan} as presented in equations \ref{fgequation} and \ref{fd}  \cite{vanilladcgan}.

\begin{equation} \label{fgequation}
	f_g(x) = \min_{z\in \mathbb{R}^{s}} \left \| x - G(z) \right \|^2
\end{equation}
\begin{equation} \label{fd}
	f_d(x) = -D(x)
\end{equation}

The paper applied another approach using Principal Component Analysis (PCA)-based novelty detection methods on the data for benchmarking. And it used Hotelling’s T2 and squared prediction error (SPE) statistics for comparison.\cite{vanilladcgan} The applied approaches in this paper were able to detect the anomaly successfully and with high accuracy.

\subsection{ Unsupervised and Semi-supervised Novelty Detection using Variational Autoencoders in Opportunistic Science Missions}

\begin{figure}
\centering
  \includegraphics[width=0.7\textwidth]{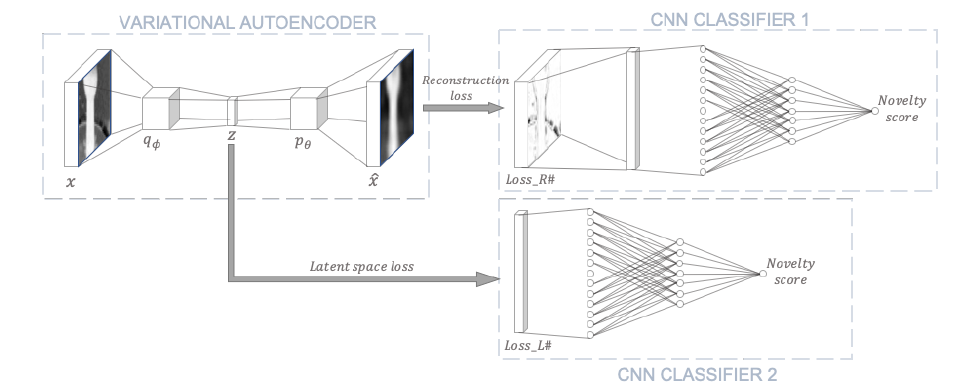}
  \caption{The structure of GAN for novelty detection approach. \cite{inspire}} \label{fig:r2structure}
\end{figure}

In the context of ensuring the safety of the robots, which are sent on scientific exploratory missions to other planets, and ensuring that the robots reach their goals and carry out the desired investigation operations without performing any unplanned actions or deviating from the desired goals, this paper provided an unsupervised and semi-supervised approach in the anomaly detection area \cite{inspire}. The approach is based on Variational Autoencoders (VAE) model and focuses on detecting the anomaly in the camera data using the data from previous scientific missions, besides providing a study about the VAE-based loss functions for generating the best reconstruction errors in detecting the anomaly features \cite{inspire}. The structure of the novelty detection approach that is used in the paper is presented in figure \ref{fig:r2structure}. Figure \ref{fig:r2structure} presented the samples generator, both the semi-supervised  model and the unsupervised model, for calculating the anomaly score.\cite{inspire}

The paper used a variety of different losses for best performance and was able to provide comparable and state-of-the-art results using the novelty score, which is the output of the applied neural networks.

\subsection{Adversarially Learned One-Class Classifier for Novelty Detection}

\begin{figure}
\centering
  \includegraphics[width=0.5\textwidth]{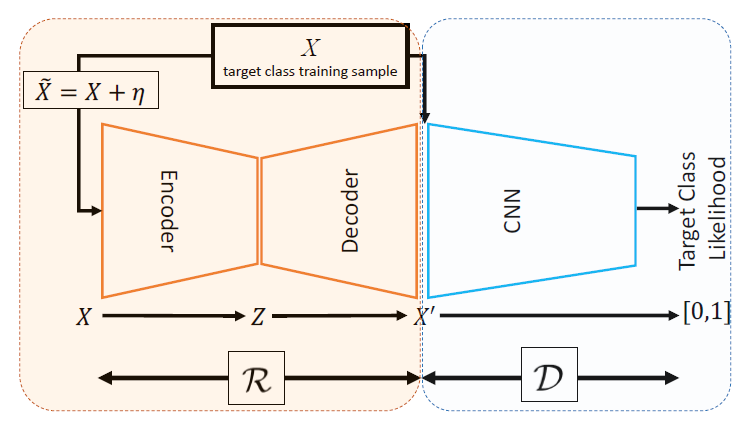}
  \caption{The Generative adversarial Network Structure. \cite{dcgan}} \label{fig:r3structure}
\end{figure}

This paper provides a one-class classification model in terms of separating the outlier samples in the data from inlier ones. The approach in the paper is based on end-to-end Generative adversarial networks in a semi-supervised manner with slight modifications in the generator\cite{dcgan}. The approach in the paper is meant to encode the typical normal images to their latent space before decoding them and reconstructing them again. The reconstructed images are sent as input to the Discriminator network for learning the normal features in those images \cite{dcgan}. Based on that, anomaly scores are given to the corresponding images to help detect whether those images are considered novel or normal, as presented in figure \ref{fig:r3structure}.

\subsection{GANomaly: Semi-Supervised Anomaly Detection via Adversarial Training} \label{sec:ganonaly}

This paper follows up on many other approaches \cite{ganomaly1} to investigate, how inverse mapping the reconstructed image to latent space is more efficient and objective than the reconstruction error between the original image and reconstructed image for anomaly detection purpose\cite{ganomaly}. The generative characteristics of the variational autoencoder give the ability to analyze the data to determine the anomaly’s cause. This approach takes into account the distribution of variables\cite{ganomaly}. \cite{SchleglSWSL17} hypothesizes that the latent space of generative adversarial networks represents the accurate distribution of the data\cite{ganomaly}. The approach proposed remapping the GAN based on the latent space. This approach \cite{ganomaly3} provided statistically and computationally ideal results by simultaneously mapping from image space to latent space\cite{ganomaly}. This paper proposes a genetic anomaly detection architecture comprising an adversarial training framework based on the previous three approaches. The approach used normal images(simple images in terms of size and the number of classes that contain) for the training (MNIST dataset and CIFAR dataset), providing excellent results.

For the purpose of our further evaluation and investigations, in this work, we  will only consider GANomaly from subsection \ref{sec:ganonaly}. We selected the approach of GANomaly since it showed very promising results in terms of accuracy on the data sets used in their work. Another interesting factor for considering their work is that they used the distance error between the latent space of  both the original image and its corresponding reconstructed output as a factor to find the anomaly images.

\section{Research questions in this work}
An anomaly in the data is considered a risk that leads to unexpected behavior in the system. It may lead to incorrect results; in some cases, it can be catastrophic and threatening. An anomaly is a deviation from the dataset distribution, an unexpected item, event, or technical glitch in the dataset that deviates from normal behavior or the standard pattern of the data. This deviation in the data may lead to the erratic behavior of the system and abnormal procedures outside the scope that the system was trained to perform. Therefore, the discovery of anomalies has become an interest to many researchers due to its significant role in solving real-world problems, such as detecting abnormal patterns in the data and giving the possibility to take prior measures to prevent wrong actions from the Camera. Inspired by the work of Rausch et al.\cite{Rausch.2021}, where research was done with the motivation to detect anomalies in the driving scene for the safety of the autonomous driving system. Their research was successfully validated and proved with the image dataset MNIST. The approach was able to reconstruct the input images using a fully-connected autoencoder network; the autoencoder network was able to learn the features of not novel images to be used later as a factor to discriminate the novelty images from the not novelty images.

In this work, we will re-evaluate the approach GANomaly, as mentioned at the end of the previous section, for the task of anomaly detection. The approach  was applied to some simple datasets MNIST and CIFAR. The approach was able to reconstruct the input images correctly and learn the features for discriminating the anomalies and not anomalies on MNIST and CIFAR. As part of our work, GANomlay will be applied to a more complex real-world driving scene dataset Berkeley DeepDrive \cite{Fisher.2020}. The complexity of the images in this dataset, such as the dimension of the image, RGB color channel, and the number of classes in each image, poses a challenge for the approach. We have formulated the contribution of our work in the following research questions (RQ) below.

\begin{itemize}

\item  \textbf{RQ.1:}\textit{ Can we reproduce the work of GANomaly on one of their used datasets?}

\item \textbf{RQ.2:} \textit{ Does such GAN approach have the ability to reconstruct high dimensional RGB  driving scene images?}

\item \textbf{RQ.3:}\textit{ Can such GAN approach be applied on highly complex driving scenes data set for task of anomaly detection?} 
\end{itemize}

\section{Evaluation process for this work}
In this section, we will explain the GANomaly architecture, the data sets, the training process, and the evaluation metric considered, as part of the evaluation process.

\subsection{GANnomaly architecture:}
\begin{figure}
\centering
  \includegraphics[width=0.5\textwidth]{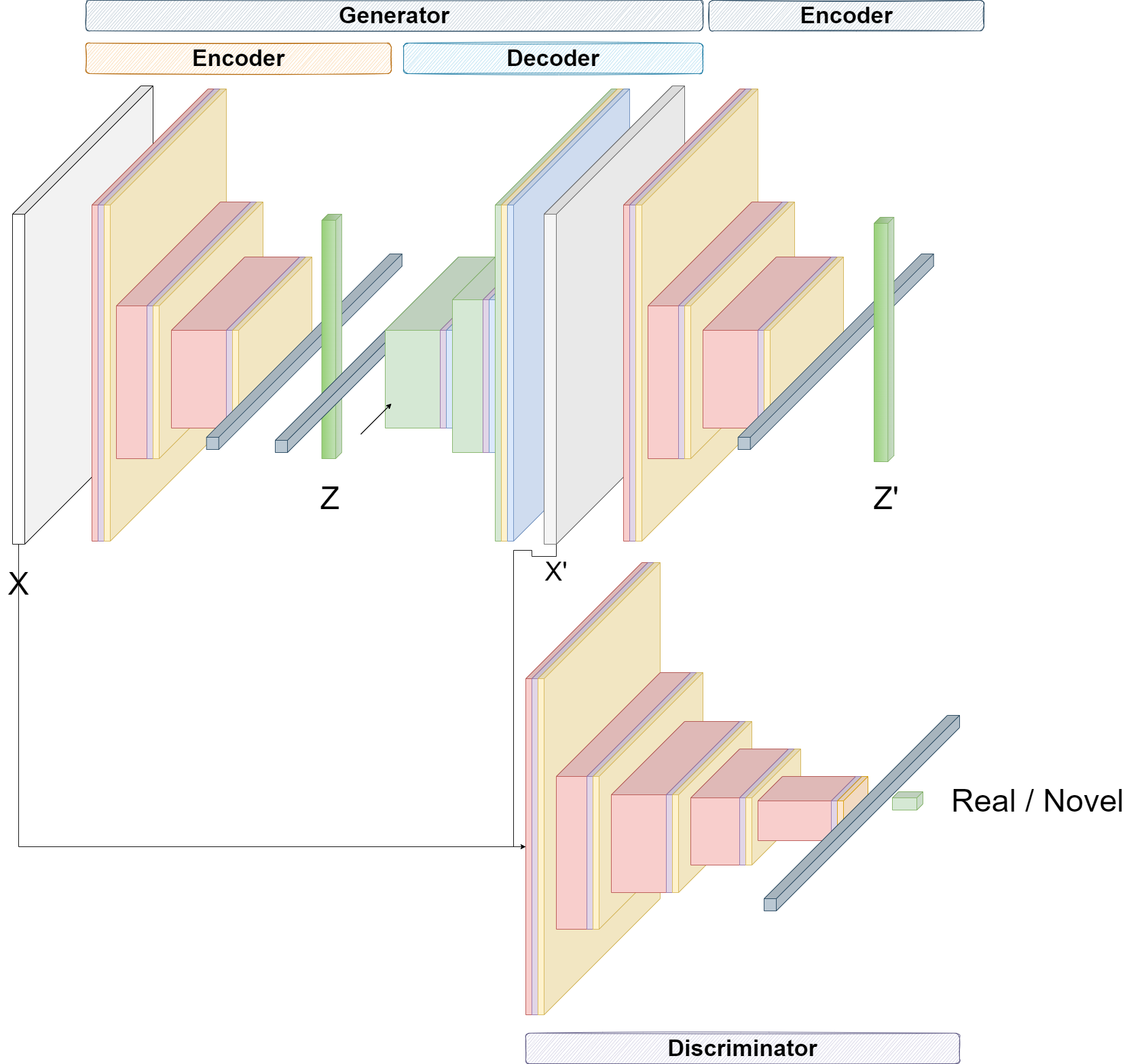}
  \caption{The structure of GANomaly \cite{ganomaly}} \label{fig:ganomaly}
\end{figure}

GANomaly is an unsupervised approach that is derived from GAN, and it was developed
for anomaly detection purposes.
The approach structure is a follow-up to the approach implemented in the paper.\cite{ganomaly}
The approach consists of three networks. The overall structure of GANomaly is illustrated
in figure \ref{fig:ganomaly}. The generator network G, which consists of encoder model $G_{E}$ and
decoder model $G_{D}$, is responsible for learning the normal data distribution which is free
of any outlier classes, and generating realistic samples. The Encoder network E 
maps the reconstructed image $\hat{X}$ to the latent space $\hat{Z}$ and finds the feature representation of the image. The discriminator network D classifies the image, whether it is real
or fake.

Generator G is an encoder-decoder model. The encoder is responsible for compressing the input sample and reducing the dimensions to vector Z (Latent space), which represents the most important features of the input sample. The decoder, on the other side, decompresses the latent space and reconstructs the input sample as realistically as possible. 

The training flow of the Generator is as follows: The Generator Encoder $G_{E}$ reads the input data X and maps it to its latent space Z, the bottleneck of the autoencoder, using three sequence groups of layers (convolutional layer, batch normalization and finally, LeakyRelu layer), downscaling the data to the smallest dimensions that should contain the best representation of the data, having the important features of the data. The Generator Decoder $G_{D}$ decodes the latent space Z and reconstructs the image again as $\hat{X}$ . $G_{D}$ acts like the architecture of DCGAN Generator, using three groups of layers (Deconvolutional layer, batch normalization, and finally, Relu layer), followed by the final layer, Tanh layer, so the values normalized between [-1, 1], upscales the vector Z and reconstructs the input image as $\hat{X}$.
The Generator reconstructs the image $\hat{X}$ = $G_{D}(Z)$ based on Latent space Z = $G_{E}(X)$.

The Encoder E acts exactly like the Generator Encoder $G_{E}$. However, E downscales the reconstructed image $\hat{X}$ to its latent space $\hat{Z}$ with a different parametrization than $G_{E}$. E learns to compress $\hat{X}$ to the smallest dimension, which should have the best representation of $\hat{X}$ but with its parametrization.
The dimension of $\hat{Z} = E(\hat{X})$ is exactly like the dimension of Z = $G_{E}(X)$ .
The Encoder is essential for the testing stage as it is part of calculating the anomaly score of the images.

The Discriminator D, which follows the architecture of DCGAN Discriminator and it is responsible for classifying the images between Fake and Normal, using five groups of layers ( (convolutional layer, batch normalization, and finally, LeakyRelu layer), followed at the end with sigmoid layer so the results would be normalized between [0, 1]. However, in the GANamly approach, anomaly detection does not rely on the Discriminator classification results. The main use of the Discriminator in this approach is for feature matching using the values in the last hidden layers before the sigmoid layer. It reduces the distance between the extracted features of the input image X and the reconstructed image $\hat{X}$ and feeds the results back to the Generator to improve its reconstruction performance. The Discriminator is trained using binary cross entropy loss with target class 1 for input images X and with target class 0 for the reconstructed images  $\hat{X}$. The loss function of the discriminator is as the following equation \ref{anoganD} \cite{ganomaly}.

	\begin{equation} \label{anoganD}
Loss_{D} = \bigtriangledown_{\theta} \frac{1}{m}\sum_{i=1}^{m}[log(D(x^{i})) + log(1-D(G(x^{i}))) ] 
	\end{equation}

The GANomaly approach hypothesizes that after compressing the abnormal image to its latent space Z = $G_{E}(X)$, the latent space would be free of any anomalies features. That is because the $G_{E}$ is only trained to compress normal images, which contain normal classes, during the training stage. As a result, the Generator Decoder $G_{D}$ would not be able to reconstruct the anomaly classes again because the developed parameterization is not suitable for reconstructing the anomaly classes. Correspondingly, the reconstructed images $\hat{X} =G_{D}(Z)$ would be free of any anomalies. 
The Encoder compresses the reconstructed image $\hat{X}$, which it hypothesized that it is free of anomaly classes, to its latent space $\hat{Z}$, which is supposed to be free from anomaly features as well.
As a result, the difference between Z and $\hat{Z}$ would increase, declaring an anomaly detection in the image.
The previous hypothesis was validated using three different types of loss functions; each of them optimizes different sub-network, and, as a final action, the total results of them would be passed to the Generator for updating its weights.

\textbf{ Adversarial Loss}: This Loss has followed the approach that is proposed by Salimans et al.\cite{salimans2016improved}; feature matching helps to reduce the instability in GANs training. Using the values of the features in the intermediate layer in the Discriminator to reduce the distance between the features representation of the input image X, that follows the data distribution $\theta$, and the reconstructed image $\hat{X}$, respectively. 
This Loss is also meant to fool the Discriminator that the reconstructed image is real. 
Let f represent the function of the output of the intermediate layer of the Discriminator. The adversarial Loss $L_{adv}$ is calculated as illustrated in the equation \ref{advLoss} \cite{ganomaly}.

	\begin{equation} \label{advLoss}
Loss_{adv} = \mathbb{E}_{X\sim \ \theta} \left \| f(X)-\mathbb{E}_{X\sim \ \theta}f(\hat{X}) \right \|_{2}
	\end{equation}

\textbf{ Contextual Loss}: This Loss is meant to improve the quality of the reconstructed image by penalizing the Generator by calculating the distance between the input image X and the reconstructed image $\hat{X}$ as the equation \ref{contLoss} \cite{ganomaly}:

	\begin{equation} \label{contLoss}
Loss_{con} = \mathbb{E}_{X\sim \ \theta} \left \| X- \hat{X} \right \|_{1}
	\end{equation}

\textbf{ Encoder Loss}: This Loss is for reducing the distance between the latent space Z that is mapped from the original image X using the Generator Encoder Z= $G_{E}(X)$ and the latent space $\hat{Z}$ that is mapped from the reconstructed image $\hat{X}$ using the Encoder $\hat{Z} = E(\hat{X})$ as the equation \ref{encodLoss} \cite{ganomaly}:

	\begin{equation} \label{encodLoss}
Loss_{enc} = \mathbb{E}_{X\sim \ \theta} \left \| G_{E}(X)- E(\hat{X}) \right \|_{2}
	\end{equation}

 The Generator learns to reconstruct images that are free of anomaly classes by both learning the Generator Encoder and the Encoder to compress normal features of the images, and they would fail to compress the abnormal features. As a result, the distance between the features of the normal image and the reconstructed image would increase and declares the anomaly in the image.
The total Loss that the Generator would depend on it for updating its weights is calculated as the equation \ref{ganomalyGLoss} \cite{ganomaly}.
	\begin{equation} \label{ganomalyGLoss}
Loss_{G} = w_{adv}Loss_{adv} + w_{con}Loss_{con} + w_{enc}Loss_{enc}
	\end{equation}

$w_{adv}$, $w_{con}$, and $w_{enc}$ are the weights parameters of the overall loss function of the Generator, which updates the Generator. The initial weights used in this approach are $w_{adv}$ = 1, $w_{con}$ = 20 and $w_{enc}$ = 1

The model is optimized using Adam optimization with a learning rate of 0.0002 with a Momentum of 0.5.

\subsection{Datasets}
 Berkeley DeepDrive (BDD) dataset includes high-resolution images (1280px x 720px) of real-live driving scenes. These images were taken in varying locations (cities streets, highways, gas stations, tunnels, residential, parking places, and villages), at three different times of the day (daytime, night, dawn), and in six different weather conditions (rainy, foggy, cloudy, snowy, clear and overcast)\cite{Fisher.2020}.
The dataset includes two packages. The first package contains 100k images. Those images include several sequences of driving scenes besides videos of those tours. The second package contains 10K images, and it is not a subset of the 100k images, but there are many overlaps between the two packages \cite{Fisher.2020}.

\begin{figure}
\centering
  \includegraphics[width=0.6\textwidth]{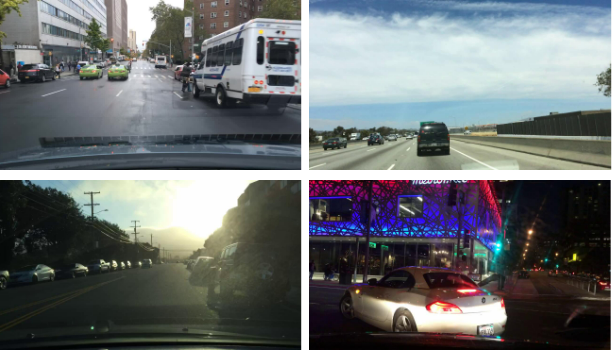}
  \caption{Samples of BDD dataset \cite{Fisher.2020}} \label{fig:bdd}
\end{figure}

BDD's usage covers many topics, including Lane detection, Road Object Detection, Semantic Segmentation, Panoptic Segmentation, and tracking.
In this thesis, a 10k package is used, and this package has two components called Things and Stuff.
The Things include countable objects such as people, flowers, birds, and animals. The Stuff includes repeating patterns such as roads, sky, buildings, and grass).
Mainly 10k package is labeled under 19 different classes of objects (road, sidewalk, building, wall, fence, pole, traffic light, traffic sign, vegetation, terrain, sky, person, rider, car, truck, bus, train, motorcycle,  bicycle)
Figure \ref{fig:bdd} illustrates some samples of the BDD 10k dataset.

Four of the 19 labels were considered novel objects for our novelty detection purpose, so the dataset is separated into two parts.
The first part, the Novel dataset, contains images that have one of the following objects listed in their labels (rider, train, motorcycle, and bicycle)
The second part, the Normal dataset, contains images that do not have any of the four novel labels mentioned lately.

\section{Evaluation}

In this section, we evaluate the performance of GANomaly based on the research questions, we formulated previously in section 4 and the evaluation process from the previous section.

\subsection{Evaluation for RQ.1:}

\begin{figure}
\centering
  \includegraphics[width=0.7\textwidth]{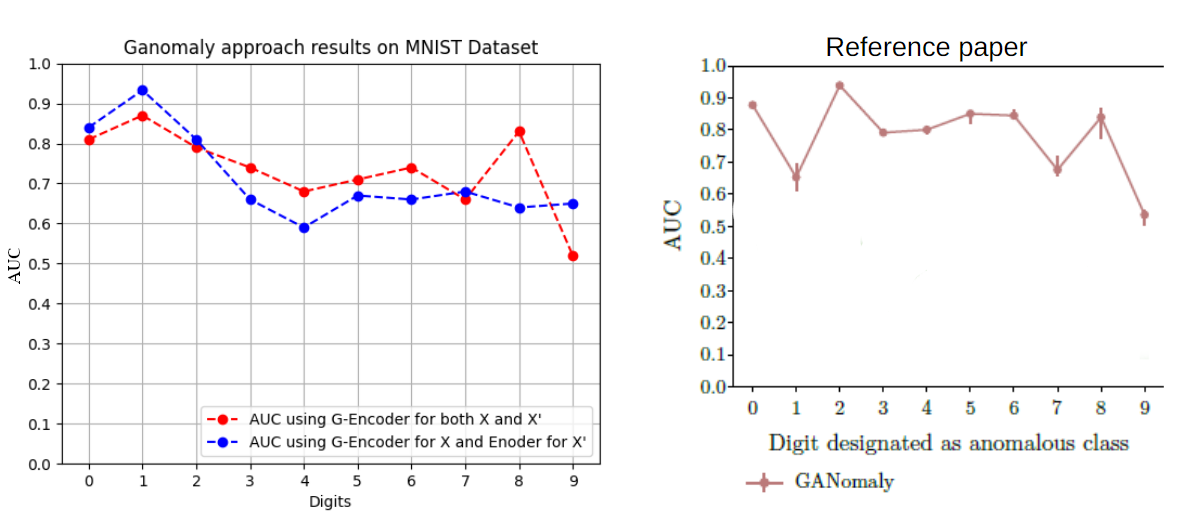}
  \caption{Left: reproduced results of the GANomaly ; right: original results in reference paper of GANomaly.\cite{ganomaly}} \label{fig:mnistauc}
\end{figure}

We replicated the results that are illustrated in the GANomaly reference paper \cite{ganomaly}, to ensure that our architecture and training method are efficient in detecting anomalies using the same dataset as the reference paper. The GANomaly setup was trained using MNIST Dataset with several anomaly parameterizations. Each time the GANomaly setup was trained by considering one digit as a novel (abnormal) while the other digits were normal. And to provide comparable results, a quantitative evaluation was applied to the results by calculating the area under the curve (AUC) for the result of each applied neural network to detect the abnormal digit.
In the GANnomaly approach that is applied in this paper, two types of anomaly scores were applied that indicate the anomalies in the reconstructed images were calculated. The original method is calculated as presented in equation \ref{ganScore1} which uses the Generator Encoder to map the input image X to the latent space Z and the Encoder which maps the reconstructed image $\hat{X}$ to the latent space $\hat{Z}$ and calculate the difference. The blue line in figure \ref{fig:mnistauc} indicates the original method. The second method presented in equation \ref{ganScore2}, gets an advantage from the Generator Encoder to map both the original image X and the reconstructed image $\hat{X}$ to their latent space Z and $\hat{Z}$ respectively. The red line in figure \ref{fig:mnistauc} indicates the second method.
Both methods are explained in detail in the section \ref{ev3subsection}

As shown in the figure \ref{fig:mnistauc}, we were able to approximate the results obtained in the reference research paper with slight differences due to modifications in parameters and hyper-parameters to get the best quality in reconstructing the complex Berkeley DeepDrive dataset, to which the approach was applied.
Figure \ref{fig:mnistauc} shows that the model achieved an excellent anomaly detection result due to the high AUC(Area under the curve) values for the digits 0,1,2 and 8. The results were better than the reference paper for digit 1,8, and they were a little less than the reference paper results for other digits 3, 4,  5, 6, and 7 and equal for some digits like 8, 9. So in reference to RQ.1, we can conclude, we were able to successfully reproduce the paper's results.

\subsection{Evaluation for RQ.2:}

\begin{figure}
\centering
  \includegraphics[width=0.6\textwidth]{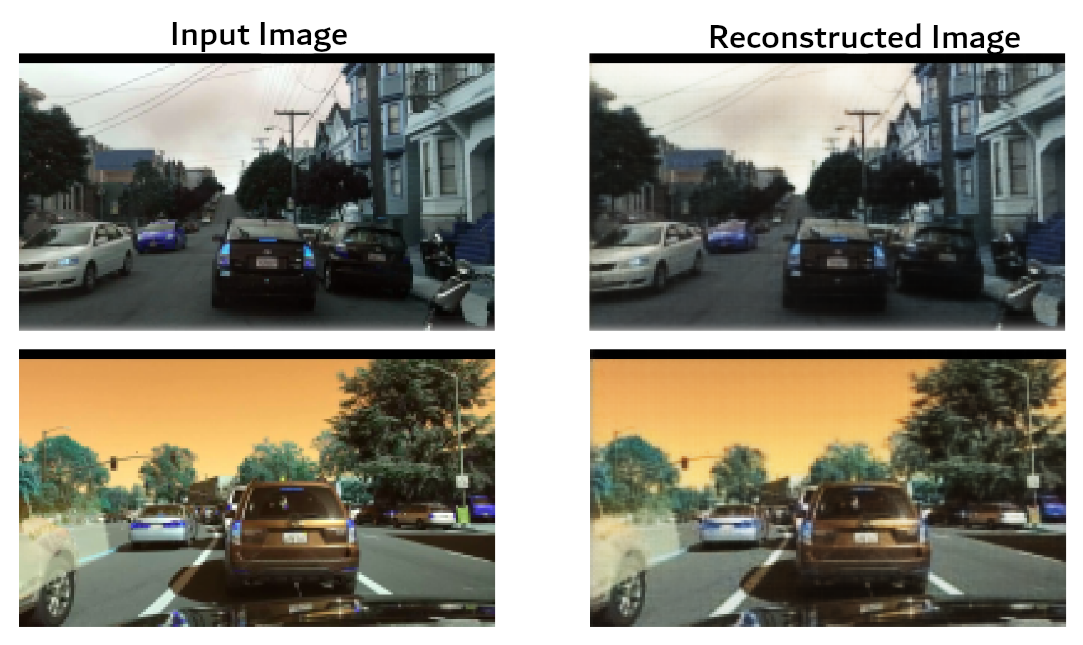}
  \caption{GANomaly performance in reconstructing the images (top with anomaly class} \label{fig:ganomalyres}
\end{figure}

During the training, only normal images were used for training with size 6239 images. For evaluation, the test sub-dataset was used, which includes 902 normal images, which are free from outlier classes, and 957 abnormal images, which contain outliers classes. The same training method mentioned on reference page \cite{ganomaly}  was followed and replicated using the MNIST dataset. With some modifications to the architecture of GANomlay, the architecture was able to reconstruct the images with high resolution with minimum reconstruction error. Figure \ref{fig:ganomalyres} illustrates the performance of the GANomaly setup in reconstructing the images.
As it is illustrated in figure \ref{fig:ganomalyres}, the top sample contains a motorcycle (abnormal class) in the bottom corner of the image. The Generator of GANomlay was still able to reconstruct the abnormal classes as efficiently as the normal classes. The region of the abnormal class was reconstructed properly without any distortion.

The unsatisfied results regarding detecting anomalies in Berkeley DeepDrive dataset can be referred to as the high complexity and the unsuitability of selected abnormal objects in the dataset.

\begin{figure}
    \centering
    \begin{subfigure}{0.31\textwidth}
        \centering
  \includegraphics[width=\textwidth]{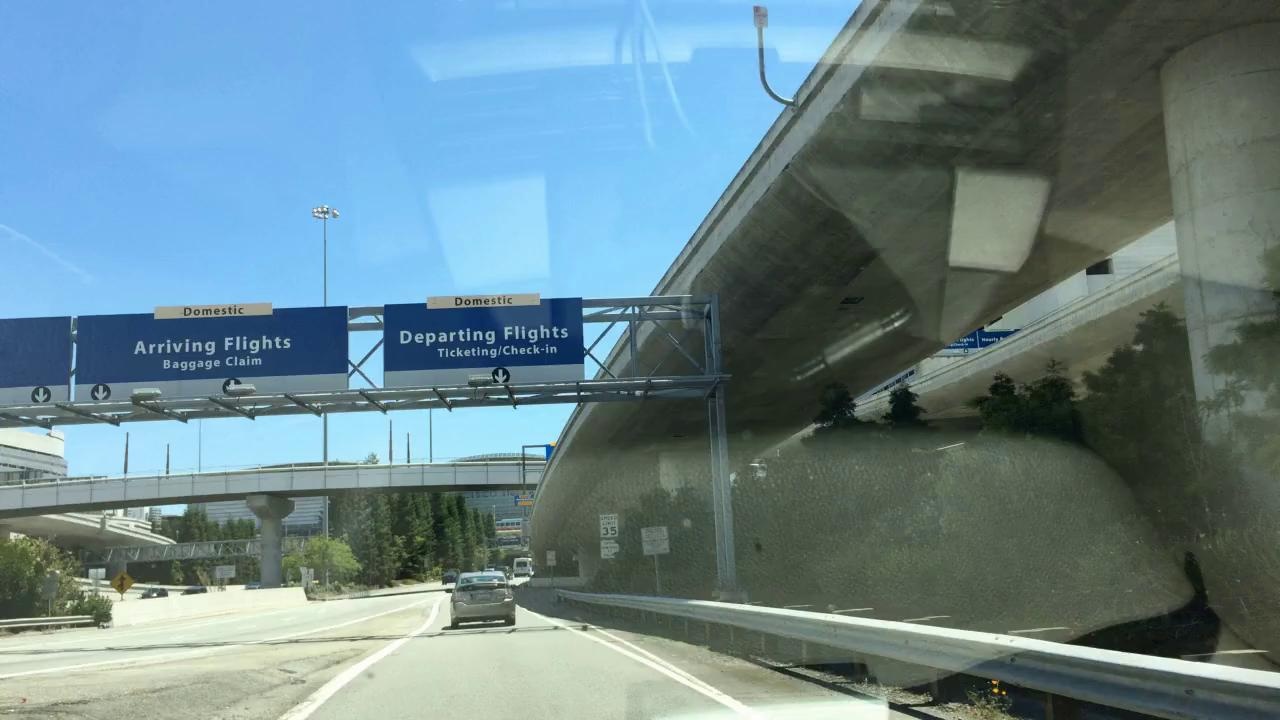}
  \caption{The image has a "Train" label but the train is not visible.} \label{fig:not1}
    \end{subfigure}
    \hfill
    \begin{subfigure}{0.31\textwidth}
        \centering
  \includegraphics[width=\textwidth]{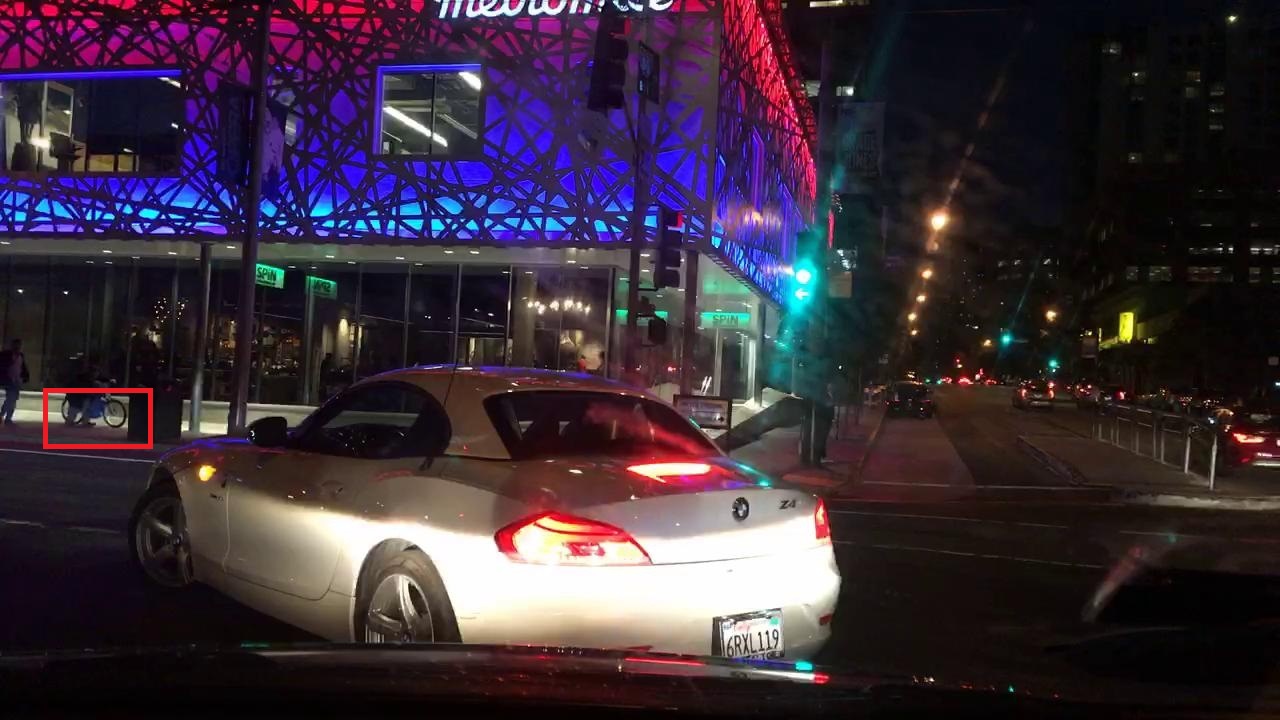}
  \caption{Abnormal classes are barely visible or small and not clear.} \label{fig:smphoto}
    \end{subfigure}
    \hfill
    \begin{subfigure}{0.31\textwidth}
        \centering
  \includegraphics[width=\textwidth]{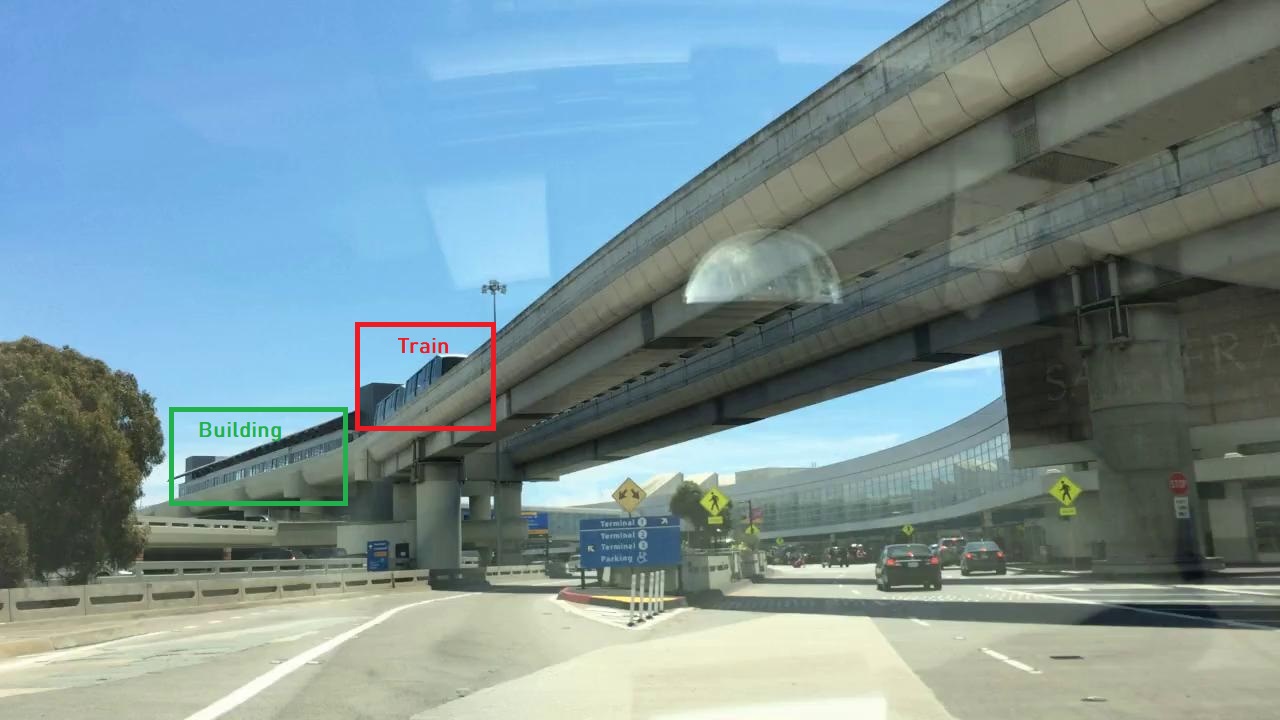}
  \caption{The train has high similarity in terms of features with a building block.} \label{fig:matching}
    \end{subfigure}
    
    \caption{Challenges and Drawback with BDD}
    \label{fig:three_figures}
\end{figure}

Analyzing the Berkeley DeepDrive dataset images show many challenges and drawback. Some of the images are labeled with some of the classes defined as abnormal in our approach. which indicate existing abnormal classes in the images, however, the abnormal classes are not visible or recognizable like in figure \ref{fig:not1} or not fully or clearly visible like in figure \ref{fig:smphoto}. In addition, some of the classes that are defined as abnormal have high similarity in terms of features with classes that are defined as normal as figure \ref{fig:matching}.

The GANomaly approach didn't succeed in detecting the abnormal classes in the reconstructed images. The Generator, despite the training process using only normal classes, was still able to reconstruct the abnormal classes during testing. This is due to the similarity in features matching between the normal and abnormal classes as mentioned previously.




So in reference to RQ.2, we could conclude that the GANomaly technique could  successfully reconstruct driving scenes of the Berkeley DeepDrive dataset.

\subsection{Evaluation for RQ.3:} \label{ev3subsection}

GANomaly is one of the newly developed approaches of GAN, and it aims to detect anomalies in the dataset rather than learning to generate samples that belong to the original data distribution. Moreover, GANomlay consists of multiple models, and it does not depend on the discriminator for discriminating the samples and classifying them into the novel and normal ones.

As mentioned in the GANomaly reference paper \cite{ganomaly}, it hypothesizes that the generator should not be able to reconstruct the outliers in the abnormal images. As a result, the Generator encoder $G_{E}$ can map the input image X to its latent space Z without the abnormal features. On the other hand, the encoder should be able to extract the full features of the image (normal and abnormal) and map it to its latent space $\hat{Z}$. So it is expected that the difference between Z and  $\hat{Z}$ will increase with increasing the outliers in the image. 
To evaluate this approach, the encoder loss $L_{enc}$  is applied to the test data $D_{test}$  by assigning an anomaly score A(x) or $S_{x}$  to the given samples x to have a set of anomaly scores as illustrated in equation \ref{ganScore1} \cite{ganomaly}.
\begin{align*}
S = \begin{Bmatrix}
s_{i}:A(x_{i}), x_{i} \in D_{test}
\end{Bmatrix}
\end{align*}

Equation \ref{ganScore1}: illustrates the anomaly score function \cite{ganomaly}.

\begin{equation} \label{ganScore1}
A(x_{i}) =  \begin{Vmatrix} G_{E}(x_{i}) - E(G(x_{i})) \end{Vmatrix} =  \begin{Vmatrix} G_{E}(x_{i}) - E(\hat{x_{i}}))) \end{Vmatrix}
\end{equation}

Another approach was applied in calculating the anomaly score. During the Generator testing, the Generator was able to reconstruct the input images successfully with its outliers but with slight distortion. 
So the Generator Encoder was applied on both the input images and the reconstructed samples, mapping them to their latent space $\hat{Z}$; as a result, we expected the abnormal features would be more extractable in the reconstructed images. The anomaly score function is transformed to the form, illustrated in equation \ref{ganScore2} \cite{ganomaly}.

 \begin{equation} \label{ganScore2}
A(x_{i}) =  \begin{Vmatrix} G_{E}(x_{i}) - G_{E}(G(x_{i}))\end{Vmatrix} \\             =  \begin{Vmatrix} G_{E}(x_{i}) - G_{E}(\hat{x_{i}}))) \end{Vmatrix}
\end{equation}

Finally, the anomaly scores in both approaches were scaled to be between 0 and 1 as the equation \ref{scaling} \cite{ganomaly}. 

	\begin{equation} \label{scaling}
\hat{s_{i}} = \frac{s_{i} - min(S)}{max(S) - min(S)}
	\end{equation}

\begin{figure}
    \centering
    
    \begin{subfigure}{0.49\textwidth}
        \centering
  \includegraphics[width=\textwidth]{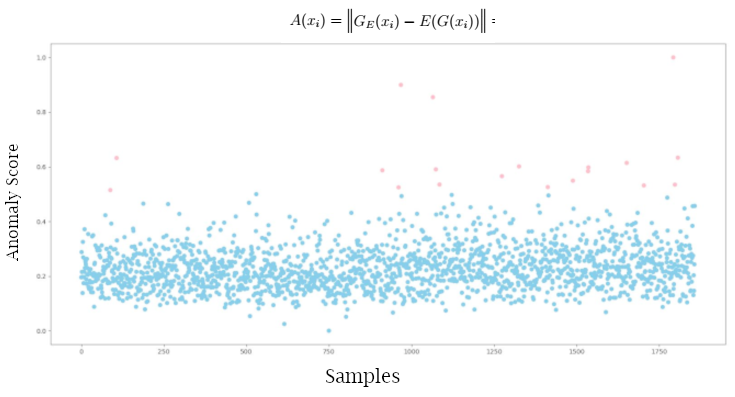}
  \caption{Scatter diagram of GANomaly scores depending on the Generator Encoder and the Encoder.} \label{fig:scatter2}
    \end{subfigure}
    \hfill
    \begin{subfigure}{0.49\textwidth}
        \centering
  \includegraphics[width=\textwidth]{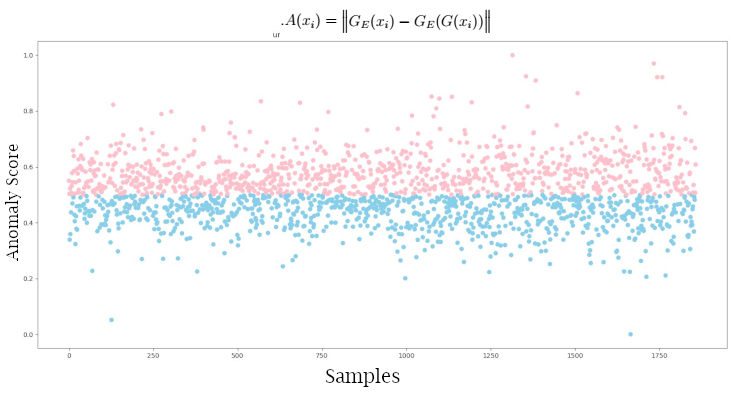}
  \caption{Scatter diagram of GANomaly scores Depending on the Encoder of the Generator only.} \label{fig:scatter1}
    \end{subfigure}
    
    \caption{Comparing results between the two approaches.}
    \label{fig:two_figures}
\end{figure}

The anomaly score was calculated for each image and then scaled to be between [0,1] depending on the max and min anomaly score between all the images. The threshold was selected by calculating the anomaly scores of the training images and the anomaly scores of 98 abnormal images, which are a subset of the abnormal images of the test images. The threshold that makes the best separation between the normal and abnormal images was selected. The evaluation was applied using a different threshold in the range [0.4, 0.6], and the threshold 0.5 made the best separation.
Figure \ref{fig:scatter2} illustrates the scatter diagram of the first approach after separation depending on threshold 0.5.


Figure \ref{fig:scatter1} illustrates the scatter diagram of the second approach after separation depending on threshold 0.5.


The confusion matrix for both GANomaly score approaches was provided to provide
a comparable result with the GANomaly reference paper and the derived metrics were
calculated. Table \ref{tab:ganrestab1} presents the confusion matrix and derived metrics
results using the Generator encoder for both the input image and the reconstructed image (we called from now GANomlay score 1) and table \ref{tab:ganrestab2} presents the confusion matrix
and derived metrics results using the Generator encoder for the input image and the encoder for
reconstructed image (we called from now GANomlay score 2). So in reference to RQ.3, we could conclude that the GANomaly technique which was successful in detecting anomalies on the MNIST dataset, was not successful when we applied this technique for anomaly detection on  the driving scenes dataset we consider in our work.

\begin{table}[h]
    \caption{Quantitative results of the two approaches.}
    \label{tab:two_tables}
    \centering
    \begin{subtable}{0.45\textwidth}
            
               \caption{A(X) = Ge(X) - E(G(X))}\label{tab:ganrestab1} \centering
\begin{tabular}{|c|c|c|c|} 
\hline
\multicolumn{2}{|c|}{\multirow{2}{*}{\begin{tabular}[c]{@{}c@{}}A(X) = Ge(X) - E(G(X))\\ Epoch 190\end{tabular}}} & \multicolumn{2}{c|}{True Values}                             \\ 
\cline{3-4}
\multicolumn{2}{|c|}{}                                                                                            & Normal                       & Novel                         \\ 
\hline
\multirow{2}{*}{\begin{tabular}[c]{@{}c@{}}Predicted\\ values\end{tabular}} & Normal                              & 886                          & 954                           \\ 
\cline{2-4}
                                                                            & Novel                               & 16                           & 3                             \\ 
\hline
\multicolumn{1}{|l|}{f1-score: 0.64}                                        & \multicolumn{1}{l|}{ACC: 0.47}      & \multicolumn{1}{l|}{P: 0.48} & \multicolumn{1}{l|}{Sn:0.98}  \\
\hline
\end{tabular}

    \end{subtable}
    \hfill
    \begin{subtable}{0.45\textwidth}
        \caption{A(X) = Ge(X) - Ge(G(X))}\label{tab:ganrestab2}\centering
\begin{tabular}{|c|c|c|c|} 
\hline
\multicolumn{2}{|c|}{\multirow{2}{*}{\begin{tabular}[c]{@{}c@{}}A(X) = Ge(X) - Ge(G(X))\\ Epoch 190\end{tabular}}} & \multicolumn{2}{c|}{True Values}                             \\ 
\cline{3-4}
\multicolumn{2}{|c|}{}                                                                                             & Normal                       & Novel                         \\ 
\hline
\multirow{2}{*}{\begin{tabular}[c]{@{}c@{}}Predicted\\ values\end{tabular}} & Normal                               & 460                          & 448                           \\ 
\cline{2-4}
                                                                            & Novel                                & 442                          & 509                           \\ 
\hline
\multicolumn{1}{|l|}{f1-score: 0.50}                                        & \multicolumn{1}{l|}{ACC: 0.52}       & \multicolumn{1}{l|}{P: 0.50} & \multicolumn{1}{l|}{Sn:0.50}  \\
\hline
\end{tabular}
    \end{subtable}
\end{table}

\subsection{Evaluation supporting RQ.3}
Another scaling approach was applied for scaling the anomaly scores for the samples. This approach depends on manipulating the threshold that separates the novel from normal images. In addition, manipulating the scaling range [Max and Min] that we depend on with scaling the anomaly scores between [0, 1]. 
This approach depends on using both the train and test data sets in figuring out the scaling ranges.
The anomaly scores for all the normal samples were scaled depending on the max and min values between the values of the anomaly scores of the normal samples as equation \ref{eq:scaling1} \cite{ganomaly}. The max and min anomaly scores for the abnormal samples were calculated as well. The anomaly scores for all the abnormal samples were scaled depending on the max and min values of the anomaly scores of the abnormal samples as equation \ref{eq:scaling2} \cite{ganomaly}.

The scaling equation for anomaly scores for normal images is:
	\begin{equation} \label{eq:scaling1}
\hat{s_{Nromal Images}} = \frac{s_{i} - min(S_{nromal})}{max(S_{nromal}) - min(S_{nromal})}
	\end{equation}

The scaling equation for anomaly scores for abnormal images is:
	\begin{equation} \label{eq:scaling2}
\hat{s_{Abnromal Images}} = \frac{s_{i} - min(S_{abnromal})}{max(S_{abnromal}) - min(S_{abnromal})}
	\end{equation}
	
The threshold was tested in the range between [0.4 to 0.55]. The threshold, that gives the best separation of the novel and normal samples, is selected.

\begin{figure}
\centering
  \includegraphics[width=0.6\textwidth]{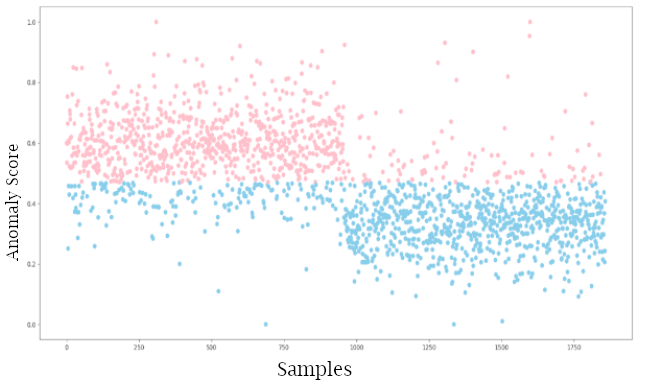}
  \caption{Scatter diagram for the scaled results using different scaling ranges. Blue points are declared as normal while pink is declared as fake images.} \label{fig:diffres}
\end{figure}

That approach provided excellent accuracy in classifying the normal image from the novel images. Table \ref{tab:gandiff} presents the confusion matrix and the accuracy of the approach using threshold 0.47 which gives the best separation of the normal and abnormal images. Figure \ref{fig:diffres} illustrates how the anomaly scores are scattered.

\begin{table}[h]
\centering
\captionsetup{skip=10pt}
\caption{Results using different scaling ranges for anomaly}
\label{tab:gandiff}
\begin{tabular}{|c|c|c|c|} 
\hline
\multicolumn{2}{|c|}{\multirow{2}{*}{\begin{tabular}[c]{@{}c@{}}Threshold = 0.47\\ acc = 96.3\%\end{tabular}}} & \multicolumn{2}{c|}{True Values}  \\ 
\cline{3-4}
\multicolumn{2}{|c|}{}                                                                                         & Normal & Novel                    \\ 
\hline
\multirow{2}{*}{\begin{tabular}[c]{@{}c@{}}Predicted \\ values\end{tabular}} & Normal                          & 878    & 44                       \\ 
\cline{2-4}
                                                                             & Novel                           & 24     & 913                      \\
\hline
\end{tabular}
\end{table}

As stated before, the fundamental idea behind anomaly detection is to detect unknown occurrences of input data not used during the training of the ML method( in our case its GANomaly), without any prior knowledge about these unknown data. In the case of GANomaly, when we used the unknown set of images to find a threshold value for classifying the input images into anomaly and not anomaly, it was able to classify most of the known images as not anomaly, and most of the unknown images as not anomaly. So based on the fundamental idea of anomaly detection, GANomaly was not able to classify the unknown driving scenes as an anomaly.

\section{Summary  and Outlook}
The GANomaly approach that is considered in this work originally demonstrated its performance of anomaly detection on image datasets MNIST and CIFAR. These image datasets contain simple images in terms of size, quality, and number of object classes in the images compared to Berkeley DeepDrive. The method was applied to RGB 32x32px images of CIFAR, as well as on the MNIST dataset, which contains gray-scale 28x28 px images. Both these data sets contain only one class per image. The evaluation of GANomaly is done based on its efficiency in detecting anomalies on driving scene images from the dataset, Berkeley DeepDrive. We used the confusion matrix to evaluate its efficiency. We were able to follow the architecture of GANomaly \cite{ganomaly} and the training method in the reference paper \cite{ganomaly} and we successfully reproduced the work of GANomaly on the MNIST dataset with our modified settings. Moreover, we were also able to reconstruct, the driving scenes input images from Berkeley DeepDrive with high resolution and quality, using this method. However, when we applied GANomaly on the  Berkeley DeepDrive for the task of anomaly detection, it suffered from low accuracy in the discriminating stage. The large number of classes in each image, the various angle of the view of each class in each image, different weather conditions, and light conditions in which images were taken, posed a great challenge for the method. In addition, the objects selected as unknown objects (trains, motorcycles, riders, bicycles) compared to known objects, occupied very fewer pixels space in the images. Another reason for the failure is, the method was able to reconstruct all the unknown objects that were not used during the training. This impacted the threshold metric used for classifying between anomaly and not anomaly images negatively. As a consequence, the threshold metric which is fundamentally based on mean squared error in the latent space, was not able to  discriminate known images( not anomaly) from unknown images(anomaly). The method classified almost all the anomaly images in the test data, as normal(not anomaly) images, hence producing a high number of false negatives. So based on the results of our GANomaly approach replication on the BDD dataset, the current state-of-the-art methods for the task of anomaly detection in such images have not provided any indication that such methods could be directly used for anomaly detection in highly complex driving scenes. With our architecture modification and hyperparameters adjustment, we were able to reconstruct the complex images in such high resolution with high quality but at the discrimination level, the conducted approach with our modification was not able to discriminate the novel objects, contained in the driving scenes. In the future, other approaches will be explored in addition to more experiments will be applied to this approach towards finding the optimal adjustments for detecting the anomalies in such highly complex driving scenes.

%
%
%
\bibliographystyle{splncs04}
\bibliography{ref}
%





\end{document}